\documentclass[runningheads]{llncs}
\usepackage{graphicx}
\usepackage{comment}
\usepackage{amsmath,amssymb} 
\usepackage{color}
\usepackage{epsfig}
\usepackage{multirow}
\usepackage{booktabs}
\usepackage{enumitem}
\usepackage{tabularx}
\newcolumntype{C}[1]{>{\centering\arraybackslash}p{#1}}
\usepackage[width=122mm,left=12mm,paperwidth=146mm,height=193mm,top=12mm,paperheight=217mm]{geometry}

\begin{document}
\pagestyle{headings}
\mainmatter
\def\ECCVSubNumber{906}  

\title{Non-Local Part-Aware Point Cloud Denoising} 

\titlerunning{Non-Local Part-Aware Point Cloud Denoising}

\author{Chao Huang\thanks{Joint first authors}
\and Ruihui Li$^{\star}$\and Xianzhi Li \and Chi-Wing Fu}
%
\authorrunning{Huang et al.}
%
\institute{The Chinese University of Hong Kong \\
\hspace{0mm}{\tt\small chaohuang1997@gmail.com}\hspace{10mm}{\tt\small \{lirh,xzli,cwfu\}@cse.cuhk.edu.hk}\qquad
}

\newcommand{\TODO}[1]{{\color{red}{[TODO: #1]}}}
\newcommand{\phil}[1]{{\color[rgb]{0.3,0.7,0.3}{#1}}}
\newcommand{\rh}[1]{{\color{black}{#1}}}
\newcommand{\hc}[1]{{\color[rgb]{1.0,0.6,0}{#1}}}
\newcommand{\xz}[1]{{\color{red}{[XZ: #1]}}}
\newcommand{\para}[1]{\vspace{.05in}\noindent\textbf{#1}}
\newcommand{\tabincell}[2]{\begin{tabular}{@{}#1@{}}#2\end{tabular}}
\def\ie{\emph{i.e.}}
\def\eg{\emph{e.g.}}
\def\etal{{\em et al.}}
\def\etc{{\em etc.}}

\maketitle

\begin{abstract}

%
This paper presents a novel non-local part-aware deep neural network to denoise point clouds by exploring the inherent non-local self-similarity in 3D objects and scenes.
%
Different from existing works that explore small local patches, we design the non-local learning unit (NLU) customized with a graph attention module to adaptively capture non-local semantically-related features over the entire point cloud.
To enhance the denoising performance, we cascade a series of NLUs to progressively distill the noise features from the noisy inputs.
Further, besides the conventional surface reconstruction loss, we formulate a semantic part loss to regularize the predictions towards the relevant parts and enable denoising in a part-aware manner.
Lastly, we performed extensive experiments to evaluate our method, both quantitatively and qualitatively, and demonstrate its superiority over the state-of-the-arts on both synthetic and real-scanned noisy inputs.



\end{abstract}

\section{Introduction}
\label{sec:intro}

The ability to remove noise in captured 3D point clouds is an indispensable tool for many applications.
With the growing availability of 3D scanning devices,~\eg, depth cameras and LiDAR sensors, 3D point clouds are becoming more common today.
However, the captured point clouds are often corrupted by noise, so point cloud denoising is essential, and typically required, as a post-processing step.
%


The goal of point cloud denoising is to recover a clean point set from a noisy input, while preserving the geometric details on the underlying object surface.
The main technical challenges to meet the goal are two-fold:
(i) how to adapt to various kinds of noise without assuming any prior on the noise distribution, and
(ii) how to maintain the data fidelity by retaining the geometric details,~\eg, sharp edges and smooth surface, in the denoised point set.
Pioneering works, such as locally optimal projection (LOP)~\cite{lipman2007parameterization} and its variants~\cite{huang2009consolidation,huang2013edge,lu2017gpf}, achieve notable successes by introducing various shape priors on the surface types and noise models to constrain the denoising process.
However, these optimization-based methods strongly rely on priors, so they may not generalize well to handle diverse inputs.
Hence, these methods tend to oversmooth or oversharpen the results, while removing the noise, particularly when the noise level is high.

\begin{figure}[!t]
\centering
\includegraphics[width=0.95\linewidth]{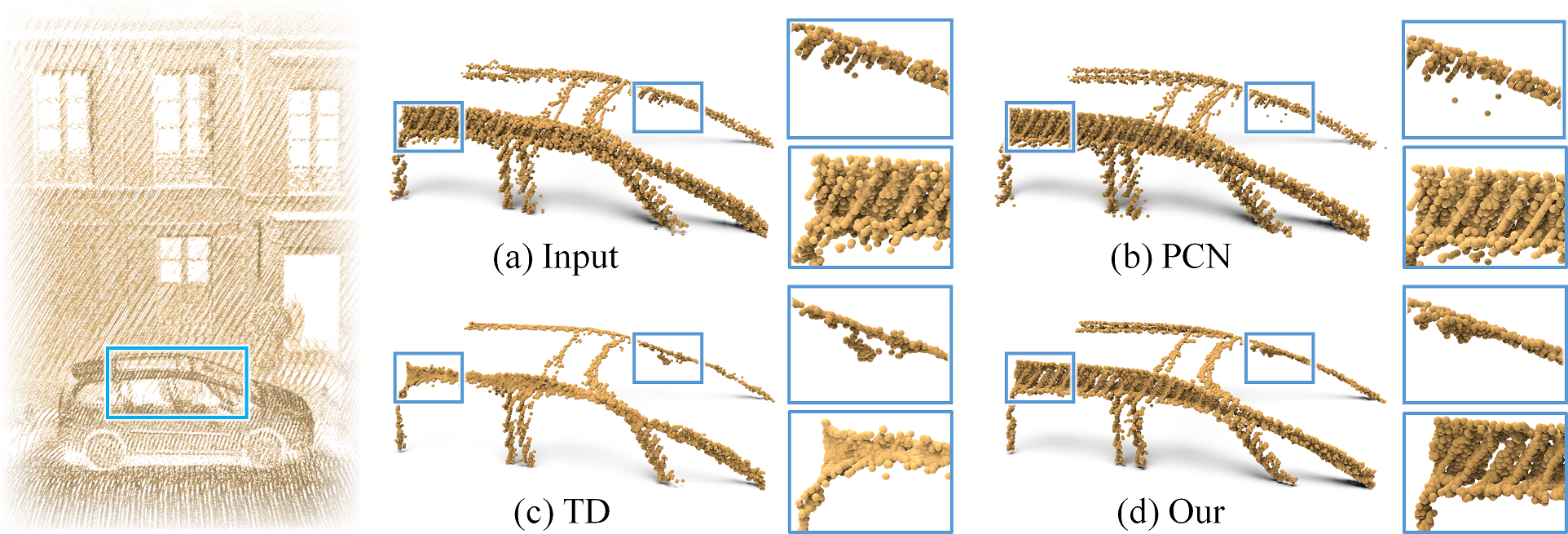}
\vspace*{-2mm}
\caption{Denoising (a) a noisy point cloud cropped from the challenging real-scanned Paris-rue-Madame dataset~\cite{serna2014paris} using
(b) PointCleanNet (PCN)~\cite{rakotosaona2019pointcleannet},
(c) TotalDenoising (TD)~\cite{hermosilla2019total}, and (d) our method.
Compared with others, our method does not produce obvious scattered points away from the object and is more faithful to the input.
}
\label{fig:teaser}
\vspace*{-8mm}
\end{figure}



Recently, many deep neural networks~\cite{qi2016pointnet,qi2017pointnet++,li2018pointcnn,zhao2019pointweb,li2019pu,zhang2019shellnet,thomas2019kpconv} are designed for analyzing point clouds.
Among them, a few recent ones aim at point cloud denoising: PointProNet~\cite{roveri2018pointpronets}, PointCleanNet~\cite{rakotosaona2019pointcleannet}, TotalDenoising~\cite{hermosilla2019total}, and DSS~\cite{yifan2019differentiable}.
These models perform denoising in 2D domain by projecting the noisy inputs into various regular representations, or directly learn a mapping from noisy inputs to noise-free outputs in a small local region.
%
Albeit the improved denoising results, their performance is typically
limited by the accuracy of the projection estimation or the local receptive field, so they may move points away from the underlying surface and fail to handle scattered points, particularly in large real scans; see Figures~\ref{fig:teaser} (b) \& (c) for results of two very recent methods.
Also, these methods are built with general network architectures.
So, they may not fully exploit problem-specific knowledge, and their models are {\em local\/} without considering the inherent non-local self-similarity in 3D objects and scenes, which has been shown to be very effective for various denoising problems~\cite{buades2005non,Schall2007,ji2010robust,ZhengSWLMCC10,gu2014weighted,zhu2017non}.

In this work, motivated by the observation that point clouds typically consist of self-similar local structures in different and possibly distant locations, we present the first attempt to formulate non-local operations in deep neural network for point cloud denoising.
Our non-local part-aware network explores points with {\em non-local semantically-related neighborhoods\/} in the input point clouds.
%
%
Specifically, we devise the {\em non-local learning unit\/} (NLU) customized with a new graph attention module, by which we enable a \emph{non-local} receptive field around each input point to effectively consider its {\em non-local semantically-related features\/} and their relations over the entire point cloud.
To enhance the denoising performance, we cascade multiple NLUs to progressively distill the noise features from the noisy inputs, while implicitly preserving the clean surface.
Lastly, to drive the network learning, besides the regular surface reconstruction loss, we further formulate a {\em semantic part loss\/} to encourage the predictions to closely locate on the relevant parts and to enable denoising in a \emph{part-aware} manner.
%
%
%

We performed various experiments to evaluate our method using synthetic and real-scanned data, and compared its performance with the state-of-the-arts, both qualitatively and quantitatively; see Figure~\ref{fig:teaser} and Section~\ref{sec:experiment}.
Experimental results show that our method is robust to handle point sets of various structures and noises, and facilitates better mesh reconstructions with finer details, as compared with others for both synthetic and real-scanned noisy inputs.

\vspace*{-4.5mm}
\section{Related Work}
\label{sec:bg}

\para{Optimization-based denoising.} \
Early methods employ various priors to reduce the error of projecting noisy inputs onto the estimated local surface.
By assuming a smooth underlying surface, Alexa~\etal~\cite{alexa2001point,alexa2003computing} introduced the moving least squares
and its robust variants~\cite{fleishman2005robust,oztireli2009feature} to preserve the features in the projection.
Later, Lipman~\etal~\cite{lipman2007parameterization} formulated the locally optimal projection (LOP) operator,
and Huang~\etal~\cite{huang2009consolidation,huang2013edge}, and
Lu~\etal~\cite{lu2017gpf} further advanced this technique,~\eg, with an anisotropic projection optimization.
These pioneering methods, however, rely on fitting local geometry,~\eg, normal estimation and smooth surface, so they may not generalize well to diverse inputs and tend to oversmooth or oversharpen the results, while removing the noise.


\para{Deep learning-based denoising.} \
A straightforward approach to adopt deep learning to the problem is to project the noisy inputs onto a 2D regular domain, then perform denoising in 2D using image-based convolutional neural networks (CNN).
Roveri~\etal~\cite{roveri2018pointpronets} designed PointProNets, a deep neural network to denoise point sets by converting them into height maps via a learned local frame.
Wang~\etal~\cite{yifan2019differentiable} formulated a differentiable surface splatting method to render noisy inputs into images and reconstruct geometric structures guided by the denoised images using the Pix2Pix~\cite{isola2017image} framework.
Their performance is, however, limited typically by the accuracy of the projection estimation.

Inspired by networks that directly process points~\cite{qi2016pointnet,qi2017pointnet++},
Rakotosaona~\etal~\cite{rakotosaona2019pointcleannet} designed PCN to predict noise-free central points from noisy patches using a point processing network.
However, since point features are extracted independently, some denoised points tend to scatter; see Figure~\ref{fig:teaser} (b).
Yu~\etal~\cite{yu2018ec} learned to detect sharp edges in small point patches and preserved edges when consolidating a point cloud; the method, however, considers mainly with small patches and requires manual edge annotations.
Zhou~\etal~\cite{zhou2019defense} adopted a statistical and non-differentiable method for removing outliers in 3D adversarial point cloud defense.
Very recently, extending Noise2Void~\cite{krull2019noise2void}, Hermosilla~\etal~\cite{hermosilla2019total} designed TotalDenoising (TD) to denoise point clouds in an unsupervised manner.
The method performs very well for flat and smooth surfaces, but for thin and fine geometric details, the results could be distorted; see Figure~\ref{fig:teaser} (c).
Different from existing works that operate locally on individual points or small patches,
we design a novel framework to capture non-local semantically-related features over the entire point cloud and progressively denoise it.

\section{Method}
\label{sec:method}




\subsection{Network Architecture Overview}
\label{subsec:network}

To start, we first introduce the basic notations and elaborate on the overall network architecture.
Given a noisy point cloud $\mathcal{P} \in \mathop{\mathbb{R}}^{N\times3}$ with $N$ points, each represented by 3D Cartesian coordinates, we aim to recover the underlying noise-free point set $\mathcal{Q} \in \mathop{\mathbb{R}}^{N\times3}$  by regressing the noise component $\mathcal{N} \in \mathop{\mathbb{R}}^{N\times3}$
of an unknown distribution, then removing $\mathcal{N}$ from input $\mathcal{P}$.
%
%

In this work, we formulate a non-local part-aware neural network to regress $\mathcal{N}$ from $\mathcal{P}$.
Overall, we have the following considerations in our approach:
%
(i) rather than working on small local patches~\cite{rakotosaona2019pointcleannet,hermosilla2019total}, we aim to distill the diverse noise in a {\em non-local\/} manner by exploring points with {\em semantically-similar features\/} but possibly far apart in space;
%
%
%
(ii) a direct shape-wise regularization tends to cause the network to focus mainly on large parts with more points and ignore small parts in shapes that are intrinsically related to the fine geometric details; so, we regularize the denoised points with respect to {\em semantically-relevant parts\/} to enable part-aware denoising; and
(iii) noise component $\mathcal{N}$ generally has small magnitude and simple patterns, so we extract the latent noise map {\em progressively\/} from the noisy inputs, instead of directly estimating the noise-free underlying geometric structures, similar to routines in image denoising~\cite{zhang2017beyond,zhang2017Learning}.

\begin{figure*}[t]
\centering
	\includegraphics[width=0.96\linewidth]{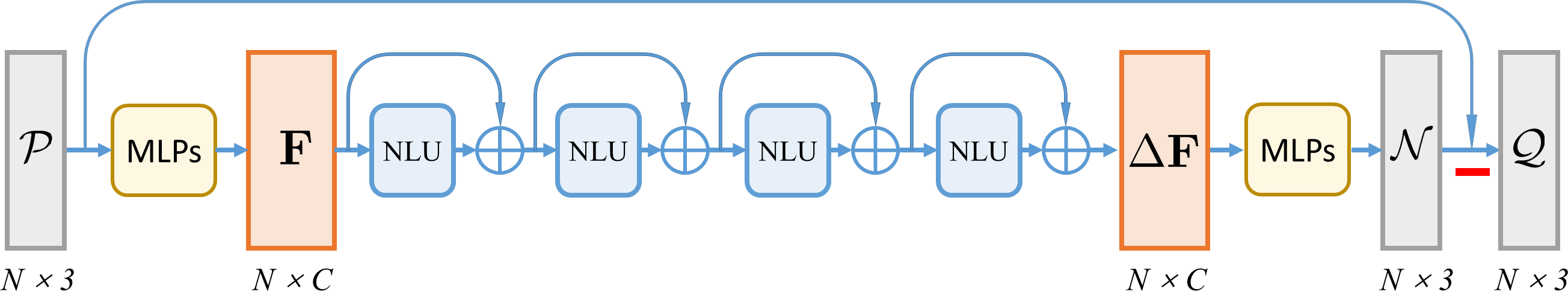}
	\caption{The overall architecture of our non-local part-aware neural network.
	$N$ is the number of points in noisy input point cloud $\mathcal{P}$, $C$ is the number of channels in feature $\mathbf{F}$, and
	NLU denotes our non-local learning unit (see Figure~\ref{fig:NLU}).
	We regress noise component $\mathcal{N}$ and remove it from $\mathcal{P}$ to produce the noisy-free output $\mathcal{Q}$.}
	\label{fig:framework}
	\vspace*{-2mm}
\end{figure*}

Figure~\ref{fig:framework} shows the overall network architecture.
Specifically, we first encode the noisy input into $C$-dimensional point-wise features $\mathbf{F} \in \mathbb{R}^{N\times C}$, which include the features of both the noise and the underlying clean surface, using a series of multi-layer perceptrons (MLPs).
Since we may not always remove noise in one shot, we thus cascade multiple residual blocks, composed by our non-local learning units (NLU) (see Section~\ref{subsubsec:NLU}) and skip to progressively distill the noise feature map $\Delta\mathbf{F}$ from $\mathbf{F}$.
%
%
In such process, an intermediate noise feature map $\Delta\mathbf{F}_i$ produced by a block is fed as input to the next block.
The initial feature $\mathbf{F}$ will go through all the blocks to produce the final noise feature $\Delta\mathbf{F}$, from which we regress noise displacement $\mathcal{N} \in \mathbb{R}^{N \times 3}$ using another set of shared MLPs.
%
%
In the end, we subtract noise displacement $\mathcal{N}$ from $\mathcal{P}$ to obtain the denoised point set $\mathcal{Q}$, and formulate a joint loss function on $\mathcal{Q}$ (see Section~\ref{subsec:loss}) to end-to-end train the whole network.
Particularly, we introduce a semantic part regularization in the loss to guide the denoising in a part-aware manner.



\subsection{Non-local Learning Unit (NLU)}
\label{subsubsec:NLU}
The purpose of the non-local learning units in the network is to distill the noise feature $\Delta\mathbf{F} \in \mathbb{R}^{N \times C}$ from the input feature $\mathbf{F} \in \mathbb{R}^{N \times C}$ by aggregating non-local semantically similar features.
Particularly, to achieve a \emph{more efficient non-local reception field\/}, we customize a novel graph attention unit inside the NLU.
By this unit, we enable adaptive aggregation of the non-local context based on the relevance from each neighboring feature to the central point being considered, rather than simply treating all the non-local similar point features equally.

\begin{figure}[!t]
	\centering
	\includegraphics[width=0.88\linewidth]{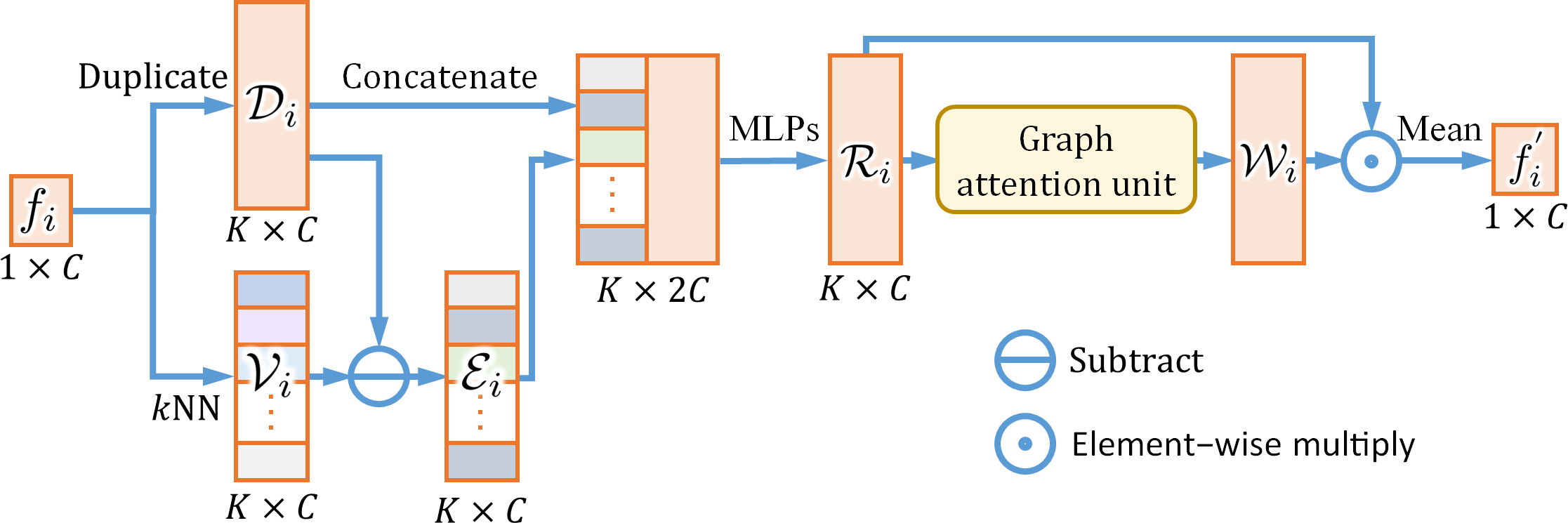}
	\vspace{-3mm}
	\caption{Illustration of our non-local learning unit (NLU). Note that we illustrate the procedure of NLU by considering the $i$-th feature vector $\mathbf{f}_i$ (for the $i$-th point) in $\mathbf{F}$.}
	\label{fig:NLU}
	\vspace{-2mm}
\end{figure}

Figure~\ref{fig:NLU} illustrates the detailed architecture of the NLU.
Note that to facilitate the understanding of the processing procedure in NLU, we present its processing procedure and architecture (see again Figure~\ref{fig:NLU}) by considering the input feature vector of the $i$-th point in the input cloud,~\ie, $f_i \in \mathbb{R}^{1 \times C}$ of $\mathbf{F}$, instead of considering the whole $\mathbf{F}$ altogether.
In the NLU, the same processing procedure (as described below) is used for all the feature vectors in $\mathbf{F}$.
%
%

Specifically, given $f_i$ in $\mathbf{F}$, we first duplicate $f_i$ $K$ times to obtain a duplicated feature map (see $\mathcal{D}_i$ in Figure~\ref{fig:NLU}) of size $K \times C$.
On the other hand, we locate $f_i$'s $K$-most similar point feature vectors by using the KNN search in $\mathbf{F}$ based on the feature similarity, and pack them together into the neighbor feature map $\mathcal{V}_i=\{f_{i1}, f_{i2}, \cdots, f_{iK}\}$ of size $K \times C$; see Figure~\ref{fig:NLU}.
%
Then, a subtraction is conducted between $\mathcal{D}_i$ and $\mathcal{V}_i$ to produce a set of edge features $\mathcal{E}_i=\{e_{i1}, e_{i2}, \cdots, e_{iK}\}$, where $e_{ij}$$=$$f_i$$-$$f_{ij}$.
We concatenate $\mathcal{E}_i$ and $\mathcal{D}_i$, followed by a series of MLPs to obtain the non-local context $\mathcal{R}_i \in \mathbb{R}^{K \times C}$ for $f_i$.
%
Note that, $\mathcal{R}_i$ encodes the \emph{non-local} similar features of $f_i$, since the KNN search is based on the feature similarity rather than the spatial distance between points.

\begin{figure}[!t]
	\centering
	\includegraphics[width=0.6\linewidth]{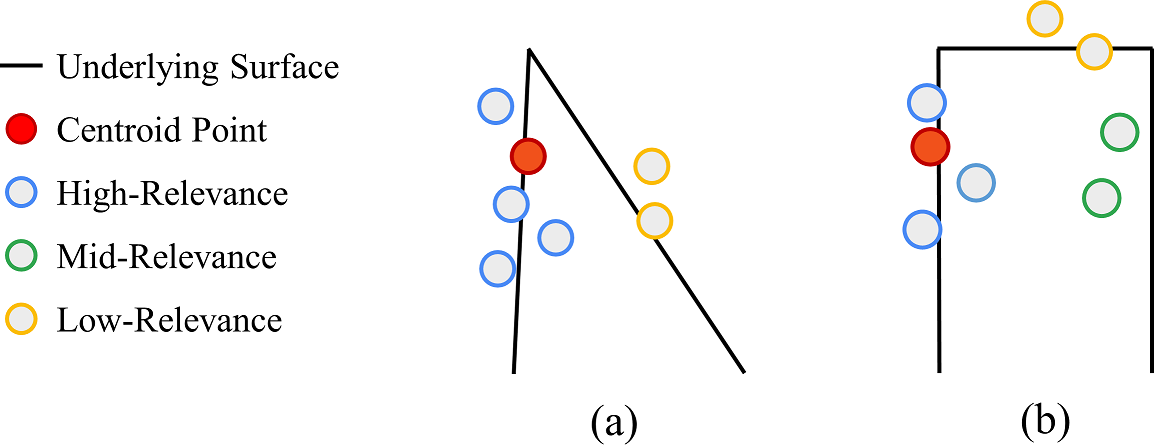}
	\vspace{-3mm}
	\caption{Degrees of relevance (similarity) between the reference point (red) and others.}
	\label{fig:Limitation}
	\vspace*{-2.5mm}
\end{figure}

Given $\mathcal{R}_i$, we may follow previous works such as~\cite{wang2018dynamic} to directly use a max-pooling or a mean operation along $K$ to produce the final extracted feature $f'_i \in \mathbb{R}^{1 \times C}$.
However, such operation equally treats all the non-local similar features, which may include dissimilar features and even outliers.
Note, for any $\mathbf{f}_i$, we may not always find $K$ semantically-relevant (similar) features in $\mathbf{F}$.
Figure~\ref{fig:Limitation} gives two illustrations, where the red points represent the reference point being considered and the other points shown in each illustration represent points with features similar (found by KNN) to the red one.
We use different colors on these points to indicate their similarity with the red point.
Clearly, not all points should contribute equally to the reference point.
%
%
In this work, we further propose a new graph attention unit to introduce a weight $\mathcal{W}_i$ to each located similar feature to enable an adaptive feature aggregation.
As shown in Figure~\ref{fig:NLU}, we obtain the attention-guided $f'_i$ by $\text{mean}(\mathcal{W}_i \odot \mathcal{R}_i)$, where $\odot$ denotes a dot product.


\begin{figure*}[t]
	\centering
	\includegraphics[width=0.9\linewidth]{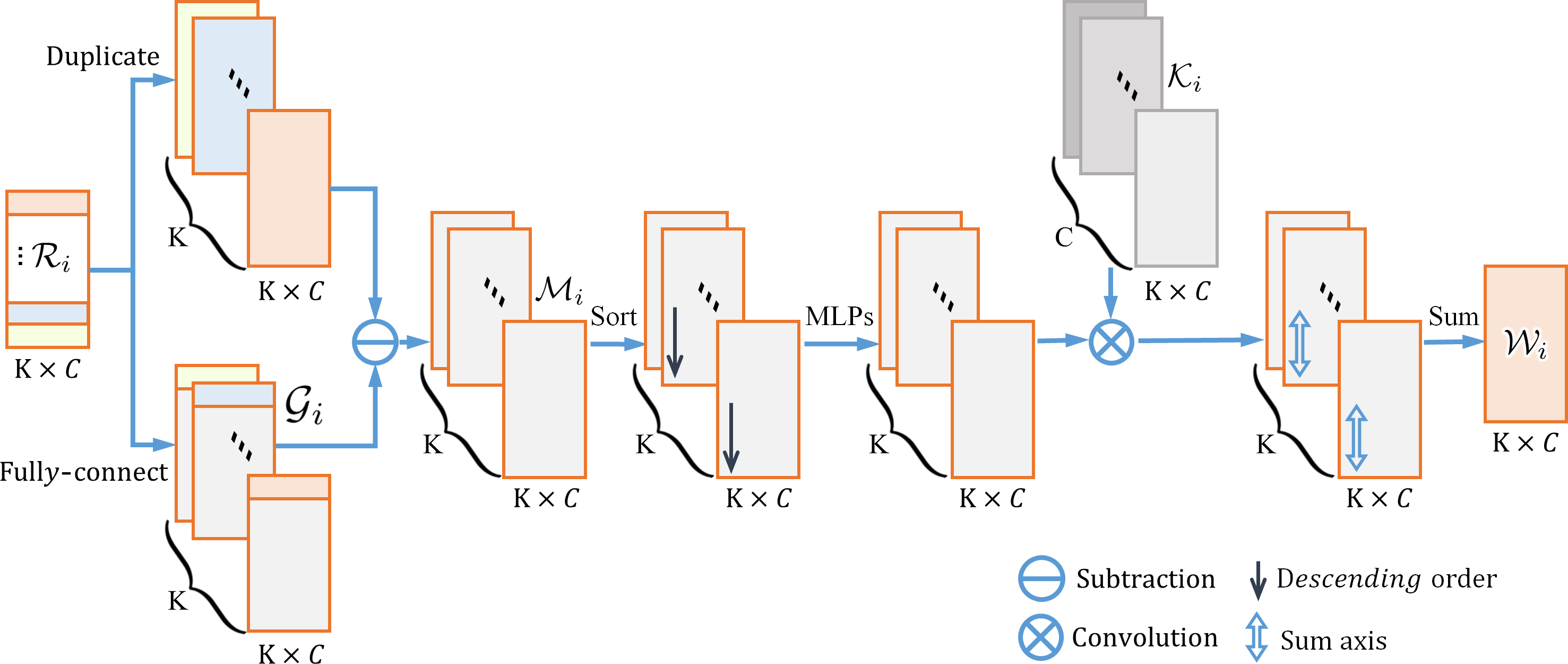}
	\caption{Illustration of the graph attention unit inside the non-local learning unit (NLU). We first calculate the reference map $\mathcal{M}_{i}$ from the input local context $\mathcal{R}_{i}$. Then we enhance it by a sorting operator and MLPs, followed by a convolution with a learnable shared kernel $\mathcal{K}_{i}$. $C$ is the number of feature channels and $K$ is the number of neighboring features.}
	\label{fig:attention}
	\vspace*{-3mm}
\end{figure*}

\para{Graph attention unit.} \
Figure~\ref{fig:attention} gives the details of our graph attention unit to regress weights for each located non-local similar feature.
Specifically, we first create $K$ copies of each row in $\mathcal{R}_i$ and pack them into a duplicated feature volume of size $K$$\times$$K$$\times$$C$.
%
On the other hand, for each row (a feature vector) in $\mathcal{R}_i \in \mathbb{R}^{K \times C}$, we treat all the other $K$$-$$1$ rows as its neighbor features, and pack them (including the row itself on the top) into one of the planes of $\mathcal{G}_i$ in Figure~\ref{fig:attention}.
Hence, $\mathcal{G}_i$ is a fully-connected feature volume of size $K$$\times$$K$$\times$$C$, which is essentially like a fully-connected graph that interconnects all pairs of features in $\mathcal{R}_i$, and each plane of $\mathcal{G}_i$ in Figure~\ref{fig:attention} corresponds to a particular row in $\mathcal{R}_i$.

%
%

Then, we subtract $\mathcal{G}_i$ from the duplicated feature volume to obtain the reference feature volume $\mathcal{M}_i \in \mathbb{R}^{K \times K \times C}$, as shown in Figure~\ref{fig:attention}.
%
Next, for each $K \times C$ feature plane of $\mathcal{M}_{i}$ (see Figure~\ref{fig:attention}), we calculate the norm of each feature vector (row) and sort the $K$ rows on each plane in descending order based on the norm value.
Hence, we can arrange the more relevant neighbor features on top.
After that, we feed the reorganized $\mathcal{M}_{i}$ into a set of shared MLPs to obtain an enhanced difference map, then perform a convolution between it and a learnable shared kernel $\mathcal{K}_{i} \in \mathop{\mathbb{R}}^{C \times K \times C}$,
followed by a sum aggregation (along the ``sum axis'' marked in Figure~\ref{fig:attention}),
to produce the final attention weight $\mathcal{W}_i$:
\begin{equation}
\mathcal{W}_i = \text{sum}(\mathcal{K}_{i} \otimes \mathcal{M}_{i}),
\end{equation}
where $\otimes$ is the convolution operation.


In previous attention methods~\cite{xie2018attentional,yang2019modeling,liu2019l2g}, they directly generalize the self-attention model~\cite{vaswani2017attention} from machine translation to handle 3D point clouds, where the attention weights are obtained by transforming the features for relation capturing.
We design the above attention unit with the goal of highlighting more relevant (or semantically more similar) neighbor features.
Hence, the attention weights are obtained based on the convolution between the sorted reference map and a learnable kernel, which is customized for the NLU and critical to improving the denoising performance; see Section~\ref{subsec:analysis} for a related analysis.



\subsection{End-to-end Training}
\label{subsec:loss}
%


Next, we present the loss function to train our network in an end-to-end fashion.
There are two considerations when we design the loss function.
\begin{itemize}[leftmargin=4.5mm]
\item
\emph{Proximity-to-surface} encourages the points in the denoised point clouds to lie on the underlying surface.
To achieve this, besides a conventional (i) shape-wise reconstruction loss to globally regularize the outputs, we further formulate a (ii) part-wise reconstruction loss to encourage the predictions towards relevant local parts to enable a part-aware denoising.
%
\vspace*{1.5mm}
\item
\emph{Distribution regularity} encourages the denoised points
to distribute more evenly on the underlying surface, instead of being cluttered together.
\end{itemize}

\noindent
Below, we present the three component terms in our loss function:

\para{(i) Shape-wise reconstruction loss.} \
Given our network output $\mathcal{Q}$ and the target point cloud $\mathcal{\hat{Q}}$, we employ the Chamfer Distance (CD) to measure the average closest point distance between them ($\|\cdot\|$ is the $L2$-norm):
\begin{eqnarray}
\label{eq:CD}
  \mathcal{L}_{shape} &=& CD(\mathcal{Q},\mathcal{\hat{Q}}) \\
    &=& \sum_{q_i \in \mathcal{Q}} \min_{\hat{q_i} \in \mathcal{\hat{Q}}}\|q_i - \hat{q_i} \| + \sum_{\hat{q_i} \in \mathcal{\hat{Q}}} \min_{q_i \in \mathcal{Q}}\|\hat{q_i}-q_i\|. \nonumber
\end{eqnarray}

\para{(ii) Part-wise reconstruction loss.} \
Shape-wise measurement alone may put more emphasis on parts with larger number of points and dismiss the smaller local parts.
As a result, denoised points may cluster in larger areas,~\eg, a table panel.
%
However, small parts in 3D shapes are usually semantically-significant, since they distinctively represent the fine geometric details in the shapes,~\eg, the ears of a cat or a lamp cable.
%
To avoid the issue, we formulate a part-wise measurement to further consider parts in the given shapes.

Specifically, benefited from the part-level annotations provided by~\cite{mo2019partnet}, we propose to calculate the CD distance (see Eq.~\eqref{eq:CD}) between parts $\mathbf{q}_i$$ \subseteq$$ \mathcal{Q}$ and $\mathbf{\hat{q}}_i $$\subseteq$$ \mathcal{\hat{Q}}$ to help recover the fine details in the denoised point cloud:
\begin{equation}
\label{eq:part}
\mathcal{L}_{part} = \sum_{i=1}^{M} CD(\mathbf{q}_i,\hat{\mathbf{q}}_i),
\end{equation}
where $M$ is the total number of parts in the object (which varies from objects to objects).
Note that, we only employ part-level annotations but not the object-level labels from the data set~\cite{mo2019partnet}.
This part-wise loss helps to emphasize the semantic parts and encourages the network predictions towards the relevant small local parts, enabling the denoising in a part-aware manner.
%
Hence, more geometric details can be preserved in our denoised points; see an experiment in the supplemental material.

\vspace*{2mm}
\para{(iii) Repulsion loss.} \
Lastly, we adopt the repulsion loss in~\cite{yu2018ec} to encourage the denoised points to move away from one another on the object surface:
\begin{equation}
\label{eq:repu}
\mathcal{L}_{repu} = \frac{1}{N \cdot k}\sum_{i=0}^N\sum_{i' \in K(i)}\eta\| q_{i'} - q_{i} \|,
\end{equation}
where $K(i)$ is the set of indices for the $k$-nearest neighbors of point $q_i \in \mathcal{Q}$, $\eta(r)$$ = $$\max(0, h^{2}$$-$$r^{2})$ indicates the penalization weights, and $h$$=$$0.03$ empirically.

\vspace*{3.5mm}
\noindent
To sum up, our joint loss function is formulated as
\begin{equation}
\mathcal{L} = \lambda_{s}\mathcal{L}_{shape} + \lambda_{p}\mathcal{L}_{part} + \lambda_{r} L_{repu},
\end{equation}
where $\lambda_{s}$, $\lambda_{p}$, and $\lambda_{r}$ are weights to balance each loss term, and we empirically set them as 0.1, 1.0, and 0.5, respectively.

\subsection{Implementation Details}

We implemented our network on PyTorch~\cite{paszke2017automatic} and trained it on a single NVidia Titan Xp GPU for 150 epochs with the Adam optimizer~\cite{kingma2014adam}.
The learning rate is initialized as $10^{-3}$ and decays by a factor of 0.2 every 10 epochs until $10^{-6}$.
In our NLU, we set $K$$=$$16$ and $C$$=$$32$.
Commonly, given a noisy point set with 20,000 points, our inference time is only around 1.5s.
Please see the supplemental material for the detailed network configuration.
We will {\em publicly release our code\/} on GitHub upon the publication of this work.

\section{Experiments}
\label{sec:experiment}

\subsection{Datasets}
\label{subsec:dataset}
\para{Training dataset.} \
%
We trained our method on the entire 3D point clouds instead of on small local patches, aiming to take advantage of the semantically-similar local regions in complete 3D shapes for point cloud denoising.
Specifically, we used the 3D benchmark dataset PartNet~\cite{mo2019partnet}, which contains not only a large amount of 3D objects, but also the semantic \rh{part} annotations.
\rh{In details,}
we randomly selected 1,600 models from the PartNet dataset~\cite{mo2019partnet} as our training dataset, covering a wide variety of 3D shapes, with simple to complex underlying geometric structures.
For each model, we directly used the 10,000 points provided in the dataset as the target noise-free output $\hat{Q}$.
We then prepared the noisy inputs by first normalizing $\hat{Q}$ to fit in a unit sphere, then corrupting the points with Gaussian noise to obtain the corresponding noisy input $\mathcal{P}$ on-the-fly.
To enrich the training samples and avoid network overfitting,
we used three levels of Gaussian noise with a standard deviation of $1.0\%$, $1.5\%$, and $2.0\%$ of the bounding box diagonal to corrupt the inputs.
In this way, we totally prepared 4,800 training pairs.
Also, we employed the provided part annotations
for calculating the part-wise reconstruction loss; see Eq.~\eqref{eq:part}.
Note that
\rh{we do not use the object-level labels, since our network does not need to know which class the point set belongs to.}
Hence, our network can generalize well to other kinds of testing shapes without additional training.

%


\para{Testing dataset.} \
We employed two kinds of testing datasets in our experiments,~\ie, synthetic and real-scanned datasets.
For the synthetic dataset, we adopted the 145 models kindly provided by~\cite{li2019pu}, which contain not only smooth structured models, but also highly-detailed models.
We uniformly sampled 20,000 points on each mesh surface as the noise-free ground truths.
Again, noise of three different levels,~\ie, 1.0\%, 1.5\%, and 2.0\%, is employed to corrupt the sampled noise-free points as the testing noisy inputs.
Note that, we directly fed the whole noisy inputs into our network to get the denoised outputs.
For the real-scanned dataset, we employed the large-scale Parisrue-Madame dataset~\cite{serna2014paris}, kindly provided by the authors.
This dataset has totally 20 millions of points.
We followed the patch-based strategies in~\cite{hermosilla2019total,rakotosaona2019pointcleannet} to first select seeds using farthest sampling, then grow a small patch of 10,000 points per seed.
Note that, these patches should cover the whole scene.
Then, we fed these patches to the network to produce the denoised outputs and combined them into the final results.



\subsection{Comparisons on Synthetic Noisy Data}
\label{subsec:comparison}

\begin{table}[t]
	\caption{Comparing our method with GPF~\cite{lu2017gpf}, PCN~\cite{rakotosaona2019pointcleannet}, and TD~\cite{hermosilla2019total} on the synthetic testing dataset. The best results are highlighted in bold. The values shown here are averaged over all the testing samples.}
	\label{tab:quan}
	\begin{center}
		\resizebox{0.8\linewidth}{!}{%
			\begin{tabular}{c||C{2.7cm}|C{1.1cm}C{1.1cm}C{1.1cm}C{1.1cm}|c}
				\toprule[1pt]
				Noise level & Metric & GPF & PCN & TD & Our & Improve on TD\\
				\hline \hline
				\multirow{5}*{1.0\%}
				&CD ($\times10^{-4}$)  & 3.99 & 2.81 & 2.22 &\bf{1.72} & 22.5\% \\   
				&HD ($\times10^{-3}$)  & 9.76 & 3.29 & 2.45 & \bf{1.63} & 33.5\%\\
				&P2F Avg ($\times10^{-3}$) & 10.26 & 5.48 & 5.51 & \bf{3.77} & 31.6\%\\
				&P2F Std ($\times10^{-3}$) & 7.84 & 4.19 & 4.41 & \bf{3.12}  & 29.3\%\\ \hline
				\multirow{5}*{1.5\%}
				&CD ($\times10^{-4}$)  & 4.16 & 3.20 & 3.13 & \bf{2.31} & 26.2\%\\ 
				&HD ($\times10^{-3}$) & 11.67 & 5.94 & 3.89 &\bf{2.74} & 29.6\%\\
				&P2F Avg ($\times10^{-3}$)  & 11.64 & 6.41 & 6.23 &\bf{5.70}& 8.5\% \\
				&P2F Std ($\times10^{-3}$)  & 8.65 & 5.49 & 5.15 & \bf{4.56}& 11.5\%  \\
				\hline
				\multirow{5}*{2.0\%}
				&CD ($\times10^{-4}$)  & 6.43 & 8.92 & 5.11 & \bf{2.65} & 48.1\% \\
				&HD ($\times10^{-3}$)  & 13.71 & 13.95 & 6.80 & \bf{3.80} & 44.1\% \\
				&P2F Avg ($\times10^{-3}$)  & 12.90 & 9.04 & 8.49 & \bf{5.91}& 31.2\% \\
				&P2F Std ($\times10^{-3}$)  & 9.69 & 8.54 & 8.07 & \bf{4.98} & 38.3\% \\
				\bottomrule[1pt]
		\end{tabular}}
	\end{center}
\vspace*{-8mm}
\end{table}

To evaluate the denoising performance of our method on synthetic noisy data, we compare our method both qualitatively and quantitatively with three state-of-the-arts, including 
GPF~\cite{lu2017gpf}, PointCleanNet (PCN)~\cite{rakotosaona2019pointcleannet}, and a recent work TotalDenoising (TD)~\cite{hermosilla2019total}.
For GPF, we used the released executable program to generate the optimal results with carefully fine-tuned parameters.
For PCN and TD, we re-trained their networks \rh{on the same synthetic dataset as ours.}


\begin{figure*}[t]
	\centering
	\includegraphics[width=0.99\linewidth]{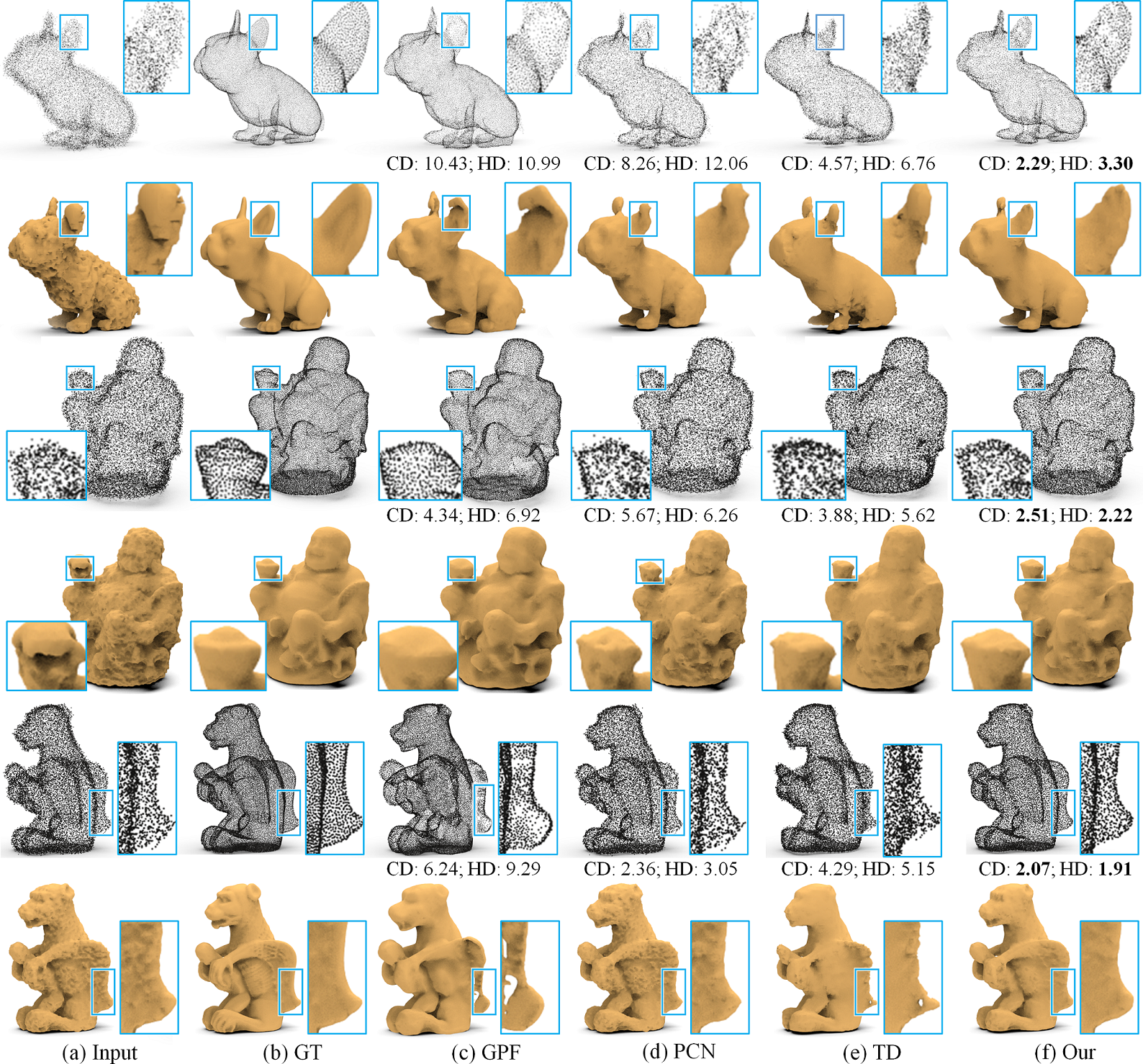}
	\caption{Comparing the denoised results produced by different methods (c-f) from the noisy inputs (a), where (b) shows the corresponding ground truths (GT).
		The three noisy inputs carry different noise levels: $2\%$, $1.5\%$, and $1\%$ (from top to bottom). \rh{The units of CD and HD are $10^{-4}$ and $10^{-3}$, respectively.} More results are shown in supp.}
	\label{fig:Synthetic}
	\vspace*{-4mm}
\end{figure*}

\para{Quantitative comparisons.} \
We first conducted a quantitative comparison on the synthetic dataset.
Since it contains ground truths, we can quantitatively measure the difference between the denoised outputs produced by various methods and the corresponding ground truths.
Here, we used four evaluation metrics:
(i) Chamfer distance (CD),
(ii) Hausdorff distance (HD),
(iii) the mean point-to-surface distance (P2F Avg), and
(iv) the corresponding standard variance (P2F Std).
The lower the metric values are, the better the denoised results are.

The evaluation results are summarized in Table~\ref{tab:quan}.
We tested each method on three levels of noise,~\ie, 1.0\%, 1.5\%, and 2.0\%.
From the results we can see that, our method clearly outperforms all the other approaches with the lowest values on all the metrics across all the noise levels.
See also the last column in Table~\ref{tab:quan}, presenting the amount of improvements achieved by our method over the recent competitive method TD.
Particularly, given inputs corrupted by a large noise level,~\eg, 2.0\%, other methods suffer from an obvious performance drop, while our method can still produce more stable denoising performance.

\begin{figure*}[t]
	\centering
	\includegraphics[width=0.95\linewidth]{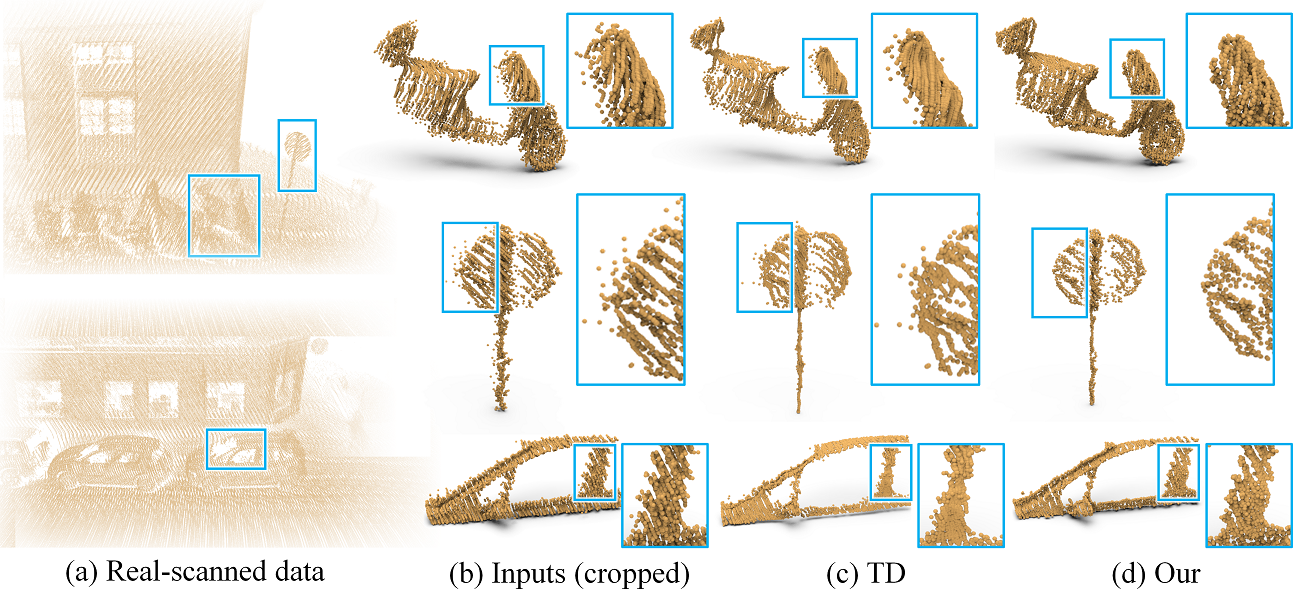}
	\caption{ Denoising (b) noisy patches cropped from (a) the real-scanned Parisrue-Madame dataset~\cite{serna2014paris} by using (c) TotalDenoising (TD)~\cite{hermosilla2019total} and (d) our method. More comparison results are shown in supplementary material.}	
	\label{fig:realworld}
	\vspace*{-4.3mm}
\end{figure*}
\para{Qualitative comparisons.} \
Besides quantitative comparisons, we further present the visual comparisons.
Figure~\ref{fig:Synthetic} shows three sets of results produced by different methods (c-f) given three different inputs (a), as well as the ground truths (b) for evaluation; see the CD and HD values below each denoised point clouds.
Note that, for each result, we present both the denoised point cloud and the corresponding surface reconstruction result row by row.
Here, we employ the Poisson surface reconstruction algorithm~\cite{poisson2006} to reconstruct the surfaces from the denoised points.
%
From the results, we can see that GPF~\cite{lu2017gpf}, which requires accurate normal estimation to guide the denoising process, is able to clean the noisy point sets for reconstructing plausible surfaces, \rh{but}
it may inflate the entire shape.
PCN~\cite{rakotosaona2019pointcleannet} can better preserve the geometric details but may generate scattered points out of the surface; see the blown-up views.
TD~\cite{hermosilla2019total} is able to recover noisy point sets and works particularly well for smooth surfaces.
However, it may oversmooth the whole surface and take away details in the results, $e.g.$, the nose of the dog (top) and the faces of the other models.
Compared with these methods, our method not only effectively removes noise and produces smooth surfaces, but also better preserves the geometric details with the lowest errors compared to the ground truths.
\rh{More comparison results can be found in the supplemental material.}

\subsection{Comparisons on Real-scanned Noisy Data}


Next, we evaluate the denoising performance of our method against the recent TD~\cite{hermosilla2019total} on real-scanned noisy data.
The comparison results are shown in Figure~\ref{fig:realworld}, where we employed two challenging large-scale scenes in Parisrue-Madame dataset~\cite{serna2014paris} as the testing inputs.
In this experiment, TD is trained on the real-scanned dataset (but without supervision), while our network is directly applied to denoise this unseen real-scanned noise pattern, after training on the synthetic dataset with Gaussian noise (see Section~\ref{subsec:dataset}).
Clearly, our method can better reveal the object's underlying surface and preserve the sharp edges and details, while TD tends to cluster the points together.
\rh{Please see the supplementary material for more comparison results with both TD and PCN.}

\newcommand{\BE}[1]{{\textbf{#1}}}
\begin{table}[t]
	\caption{ Comparing the denoising results in terms of CD ($\times10^{-4}$) between our method and the state-of-the-arts on a synthetic testing dataset with unseen noise levels.}
	\vspace{-2mm}
	\label{tab:robustness}
	\begin{center}
	\resizebox{0.75\linewidth}{!}{%
	\begin{tabular}{C{1cm}|C{1cm}C{1cm}C{1cm}|C{2cm}C{2cm}C{2cm}}
	\toprule[1pt]

	& \multicolumn{3}{c|}{singe noise level} & \multicolumn{3}{c}{mixed noise level}\\
				\cline{2-7}
				& $2.5\%$ & $3.0\%$ & $3.5\%$ &$1.0\% \& 2.0\%$ & $1.0\% \& 3.0\%$ &$2.0\% \& 3.0\%$ \\
\hline
                PCN & 7.46 & 8.59 & 10.82 & 6.78 & 7.47 & 9.28 \\
				TD & 6.82 & 7.06 & 9.96 & 6.21 & 6.99 & 7.11 \\ \hline
				our & \BE{4.39} & \BE{5.96}& \BE{8.08}& \BE{1.90} & \BE{2.75} & \BE{4.28} \\
\bottomrule[1pt]
\end{tabular}}
\end{center}
\vspace*{-4mm}
\end{table}

%

\begin{figure*}[!t]
	\centering
	\includegraphics[width=0.95\linewidth]{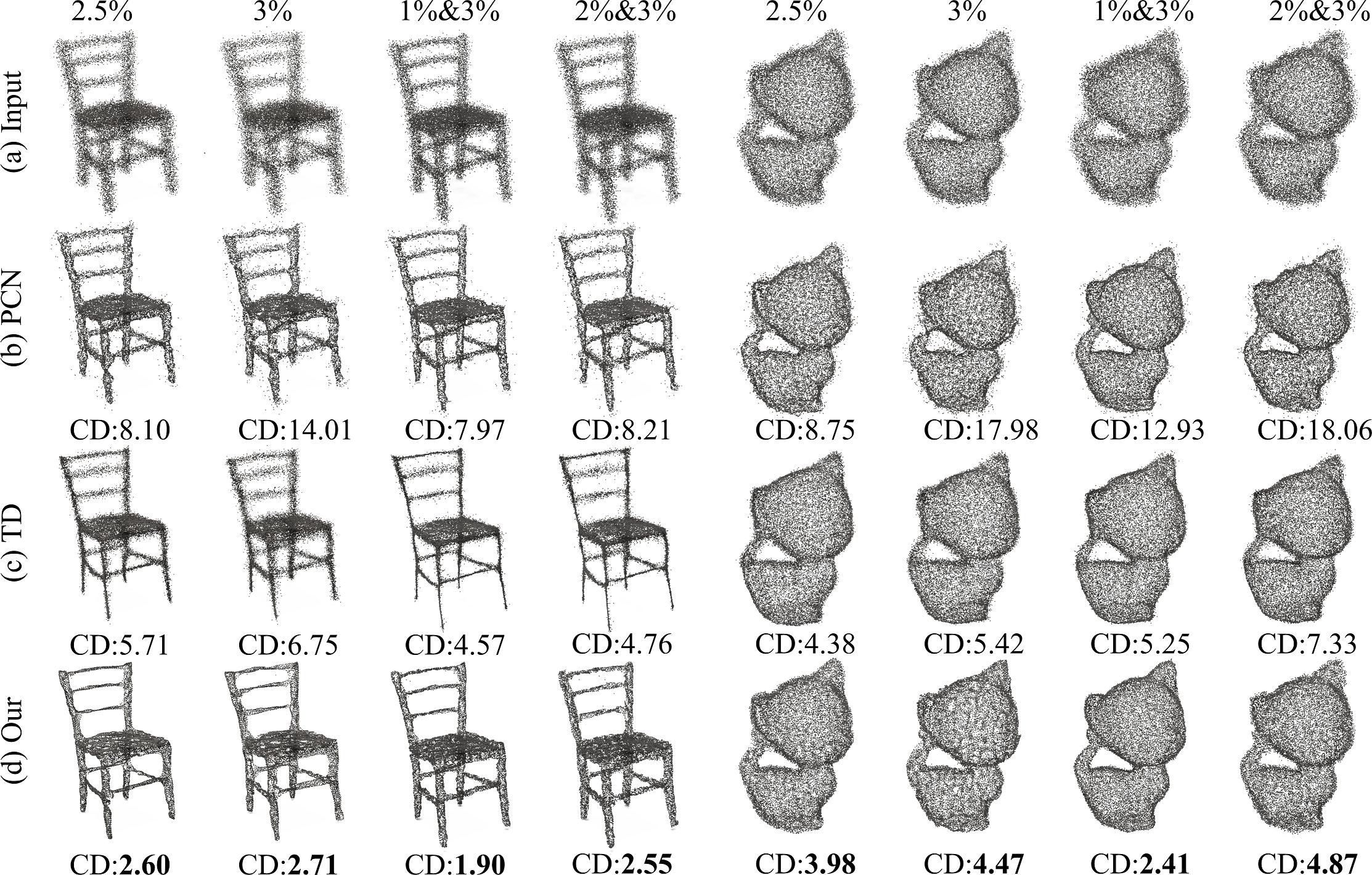}
	\caption{Comparing the denoising results by applying PCN (b), TD (c), and our method (d) to inputs (a) with unseen noise levels. More comparison results are shown in supp.}
    \label{fig:noiselevels}
	\vspace*{-3mm}
\end{figure*}


%


\begin{table}[t]
\caption{Ablation study on major components in our method. $\text{NLU}^*$ represents the NLU without the graph attention unit.}
\vspace{-2mm}
\label{tab:ablation}
\begin{center}
	\resizebox{0.57\linewidth}{!}{%
		\begin{tabular}{l|ccccc|c}
		\toprule[1pt]\noalign{\smallskip}
		Model & $\mathcal{L}_{shape}$  & $\mathcal{L}_{part}$ & $\mathcal{L}_{repu}$ & $\text{NLU}^*$ & NLU & CD ($\times 10^{-4}$)\\
		\hline
				A & $\checkmark$  &  &  & & & 2.76  \\
				B & $\checkmark$  & $\checkmark$  &  & & & 2.62 \\
				C & $\checkmark$  & $\checkmark$  & $\checkmark$ & & & 2.60  \\
				D & $\checkmark$  & $\checkmark$  & $\checkmark$  & \checkmark & & 2.48 \\
				E (full) & $\checkmark$  & $\checkmark$  & $\checkmark$  &   & \checkmark  &\textbf{2.31} \\
\bottomrule[1pt]
\end{tabular}}
\end{center}
\vspace*{-9mm}
\end{table}


\subsection{Model Analysis}
\label{subsec:analysis}

\para{Noise Tolerance Test.}
To analyze the noise tolerance of our network, we employed our previously-trained network on a synthetic dataset, which is corrupted by different levels of Gaussian noise: 1.0\%, 1.5\%, and 2.0\%, as well as test noisy models with unseen large noise levels and even mixed noise levels.
Table~\ref{tab:robustness} reports the comparison results in terms of the CD metric.
Note that, the shown values are averaged over all the testing models of the same category (noise level).
Clearly, our method consistently outperforms others across all the noise settings.
Also, we provide the visual comparisons in Figure~\ref{fig:noiselevels}.
Compared with our results (bottom row), other methods tend to retain excessive noise.
\rh{Please see the supplementary material for more comparison results.}

\para{Ablation study.}
To analyze the contribution of major components in our method, including the shape-wise reconstruction loss, part-wise reconstruction loss, repulsion loss, NLU module and graph attention unit, we conducted an incremental ablation study.
Specifically, we first prepared a baseline model (denoted as A) by only keeping the shape-wise reconstruction loss.
Based on model A, we then added back the remaining components back one at a time to obtain a new model, which is denoted as B to E, respectively; see Table~\ref{tab:ablation} for the detailed configuration of each model.
Naturally, model E is our full pipeline with all the major components.
We then trained and tested each model on the synthetic dataset with a noise level of 1.5\%.

Table~\ref{tab:ablation} summarizes the evaluation results.
Note that, for these models without NLU,~\ie, models A, B, and C, we employ MLPs for feature extraction.
Clearly, model E (our full pipeline) achieves the best denoising performance with the lowest CD value, showing that each component contributes to improve the denoising performance.
Particularly, comparing the CD values achieved by model B and model A we can see that, adding the part-wise reconstruction loss back leads to a notable performance improvement.
Similarly, adding our NLU back also improves the performance obviously; see the comparisons between model C and model D.
The two observations clearly demonstrate the effectiveness of the non-local part-aware context learning approach for point cloud denoising.

\para{Analysis on the number of NLUs.}
In our default settings, we cascade four NLUs; see Figure~\ref{fig:framework}.
To explore the effect of using different number of NLUs, we further re-implemented our network with three and five NLUs, respectively, and trained each network on the same synthetic dataset with a noise level of 1.5\%.
The testing results show that, the CD values ($\times 10^{-4}$) of using three, four (default), and five NLUs are 2.40, 2.31, and 2.30, respectively.
Although the performance of using five NLUs is slightly better than our default setting, the network has 55.7K parameters as compared to 44.6K parameters with four NLUs.
Hence, we choose to use four NLUs to balance the performance and efficiency.
%

\begin{figure*}[t]
	\centering
	\includegraphics[width=0.75\linewidth]{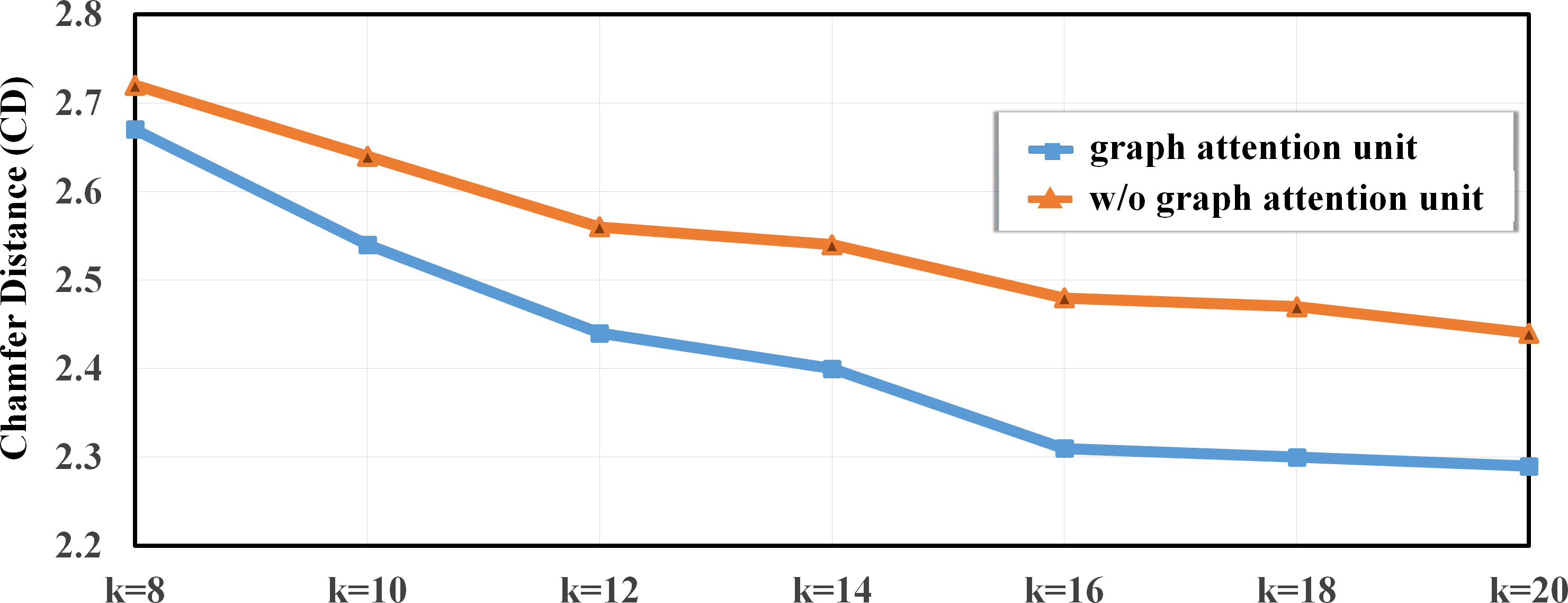}
	\caption{Denoised performance by increasing the hyper-parameter $K$ in
our NLU with or without the graph attention unit. Note that the unit of CD  is $10^{-4}$.}
    \label{fig:varyingk}
	\vspace*{-5mm}
\end{figure*}
\para{Analysis on the hyper-parameter $K$ in NLU.} \
Generally, a smaller $K$ in NLU means that we take fewer neighboring features into the local context, resulting in a smaller receptive field.
However, a larger $K$ may involve some unrelated features of different underlying geometric structures.
Our graph attention unit is to learn the weights among the neighbor point features, therefore enabling our method to attend to more relevant features.
Hence, our method benefits more from a larger $K$ and achieves a better denoising performance.
To verify this, we test the denoising performance for different choices of $K$ with/without our graph attention unit, and the results are summarized in Figure~\ref{fig:varyingk}.
Clearly, when we remove the graph attention unit, the CD values are much higher across all the $K$ values; see the orange plot vs. blue plot.
Further, as shown in the blue plot, the denoising performance keeps improving as $K$ increases, and finally converges at around $K$$=$$20$.
In our work, we choose $K$$=$$16$ by considering both the denoising performance and computational efficiency.

\section{Conclusion}
\label{sec:conclusion}


In this paper, we presented a non-local part-aware deep neural network to progressively denoise point clouds.
%
Different from existing works that perform denoising in small local regions, we consider the inherent non-local self-similarity in 3D objects and scenes, by designing the NLU customized with a graph attention module to explore semantically-relevant features in point clouds non-locally.
Also, to preserve more geometric details in the denoised point clouds, besides the regular surface reconstruction loss, we empower our network with part-awareness by formulating a semantic part loss to encourage the predictions to closely locate on the relevant parts.
Extensive experiments demonstrated the effectiveness of our method, showing that it outperforms the state-of-the-arts in various configurations, from synthetic data set to large real-scanned point clouds.

Since our method is trained in a supervised manner, our approach typically requires the provision of ground truth information in the training.
In the future, we plan to explore a weakly-supervised framework to simultaneously learn from both the synthetic and real-scanned inputs, by means of various domain adaptation strategies.
We will also consider the potential of multi-task learning, by performing denoising with upsampling
together, to let
the network learn to produce dense and cleaned outputs, to address the issues of both sparsity and noise, which are particularly common in real-scanned data.

%
%
%

%

%
%
\bibliographystyle{splncs04}
\bibliography{egbib}
\end{document}